\documentclass[runningheads]{llncs}
\usepackage{graphicx}

\usepackage{multirow, makecell}
\usepackage{xcolor}
\usepackage{graphicx}
\usepackage{enumitem}
\usepackage{ulem}

\usepackage{times}
\usepackage{epsfig}
\usepackage{graphicx}
\usepackage{amsmath}
\usepackage{amssymb}
\usepackage{wrapfig}

\usepackage{hyperref}

\begin{document}
\title{ASC-Net: Adversarial-based Selective Network for Unsupervised Anomaly Segmentation}
\titlerunning{ASC-Net}
%
\author{Raunak Dey \inst{1}\and Yi Hong\inst{2}}
 

%

%
\institute{Department of Computer Science, University of Georgia\\ \email{raunak.dey@gmail.com} \and
Department of Computer Science and Engineering, Shanghai Jiao Tong University \\ \email{yi.hong@sjtu.edu.cn}}

%
\maketitle              

\begin{abstract}
We introduce a neural network framework, utilizing adversarial learning to partition an image into two cuts, with one cut falling into a reference distribution provided by the user. This concept tackles the task of unsupervised anomaly segmentation, which has attracted increasing attention in recent years due to their broad applications in tasks with unlabelled data. This Adversarial-based Selective Cutting network (ASC-Net) bridges the two domains of cluster-based deep learning methods and adversarial-based anomaly/novelty detection algorithms. We evaluate this unsupervised learning model on BraTS brain tumor segmentation, LiTS liver lesion segmentation, and MS-SEG2015 segmentation tasks. Compared to existing methods like the AnoGAN family, our model demonstrates tremendous performance gains in unsupervised anomaly segmentation tasks. Although there is still room to further improve performance compared to supervised learning algorithms, the promising experimental results shed light on building an unsupervised learning algorithm using user-defined knowledge.


\end{abstract}

\section{Introduction}
\label{section:intro}
In computer vision and medical image analysis, unsupervised image segmentation has been an active research topic for decades~\cite{giordana1997estimation,lee2002unsupervised,o2004combined,puzicha1999histogram,shi2000normalized}, due to its potential of applying to many applications without requiring the data to be manually labelled. Recently, advances in GANs \cite{goodfellow2014generative} have given rise to a class of anomaly detection algorithms, which are inspired by AnoGAN \cite{schlegl2017unsupervised} to identify abnormal events, behaviors, or regions in images or videos~\cite{del2016discriminative,erfani2016high,seebock2016identifying}. The AnoGAN learns a manifold of normal images by mapping from image space to a latent space based on GANs. To detect the anomaly, AnoGAN needs iterative search in the latent space to find the closest corresponding images for a query image. The AnoGAN family, including f-AnoGAN~\cite{schlegl2019f} and other works~\cite{baur2018deep,berg2019unsupervised,kimura2018anomaly,zenati2018efficient,zenati2018adversarially}, focus on the reconstruction of the corresponding normal images for a query image, but not directly working on the anomaly detection. As a result, their reconstruction quality heavily affects the performance of anomaly detection.

\begin{figure}[t]
\centering
\includegraphics[width=0.8\columnwidth]{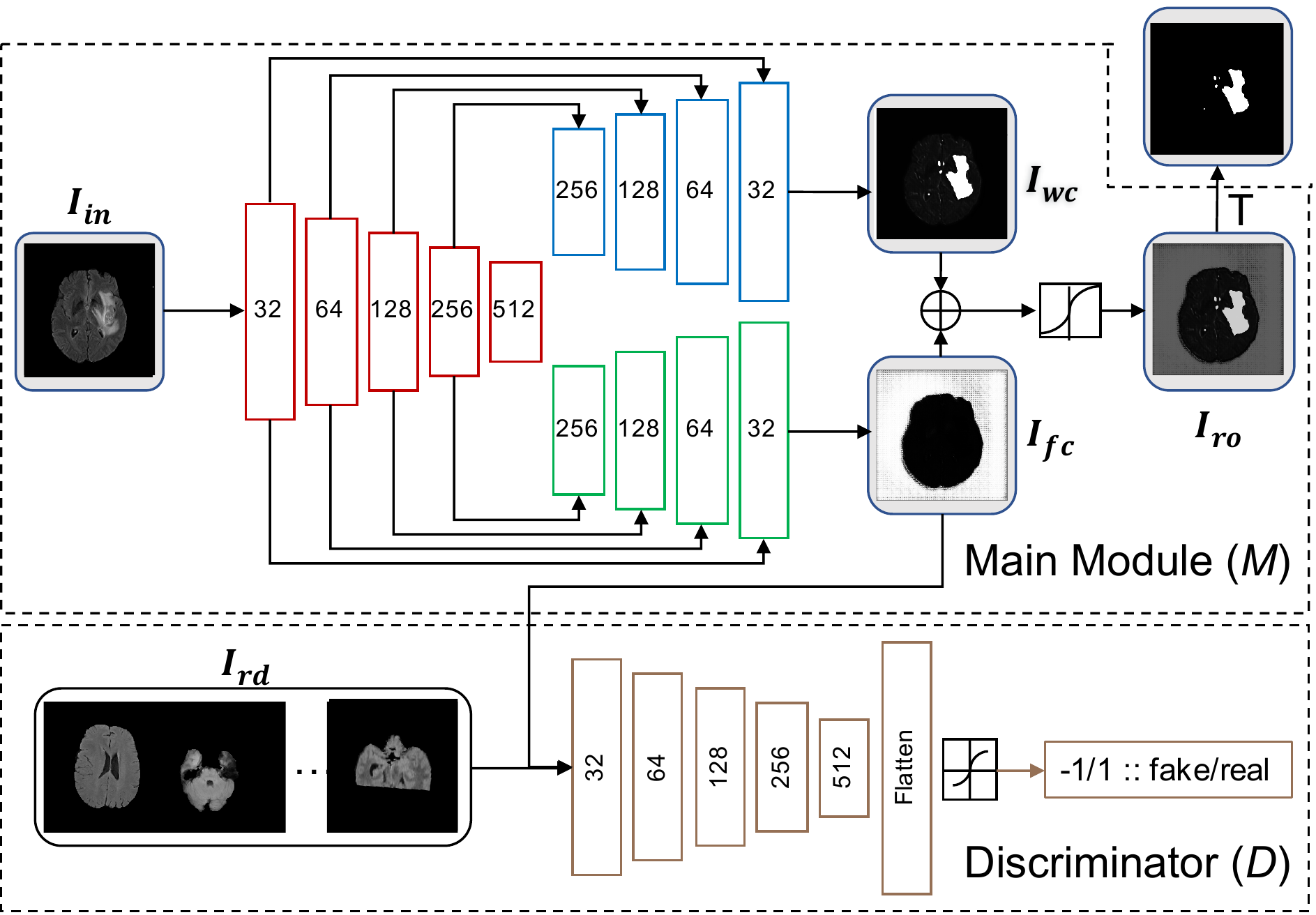}
\caption{Overview of our proposed ASC-Net for unsupervised anomaly segmentation.}
\label{fig:overview}
\end{figure}

To center the focus on the anomaly without needing faithful reconstruction, we propose an adversarial-based selective cutting neural network (ASC-Net)\footnote{Our source code is available on Github: \url{https://github.com/raun1/ASC-NET}.}, shown in Figure~\ref{fig:overview}. This network aims to decompose an image into two selective cuts based on a reference image distribution. Typically, the reference distribution is defined by a set of images provided by users or experts who have vague knowledge and expectation of normal cases. In this way, one cut will fall into the reference distribution, while other image content outside of the reference image distribution will group into the other cut. These two cuts allow to reconstruct the original input image semantically and perform a simple intensity thresholding to cluster normal and abnormal regions. To consider these two cuts simultaneously, we extend U-Net~\cite{ronneberger2015u}  with two upsampling branches, as used in CompNet~\cite{dey2018compnet}, a supervised image segmentation approach. Meanwhile, one branch connects to a GAN's discriminator network, which allows introducing the knowledge contained in the reference image distribution. With the discriminator component aiding, the network can separate images into softly disjoint regions; that is, the generation of our selective cuts is under the constraint of the reference image distribution. As a result, we obtain a joint estimation of anomaly and the corresponding normal image, thus bypassing the need for perfect reconstruction. Furthermore, under the constraints of the GAN discriminator and the reconstruction of the original input, our ASC-Net becomes an unsupervised solution for anomaly detection, since we do not have any labels for the anomaly, with only a collection of normal images in the reference distribution.

We evaluate our proposed unsupervised anomaly segmentation network on three public datasets, i.e., MS-SEG2015~\cite{carass2017longitudinal}, BraTS-2019~\cite{bakas2017advancing,bakas2018identifying,menze2014multimodal}, and LiTS~\cite{bilic2019liver} datasets. For the MS-SEG2015 dataset, an exhaustive study on comparing multiple existing autoencoder-based models, variational-autoencoder-based models, and GAN-based models is performed in~\cite{baur2021autoencoders}.
Compared to the best Dice scores reported in~\cite{baur2021autoencoders}, we have significant gains in performance, which are increased by 23.24\% without post-processing and 20.40\% with post-processing\footnote{Different from that in~\cite{baur2021autoencoders}, we use a simple open-and-closed operation for post-processing.}.
For BraTS dataset, our experiments show that f-AnoGAN, the one performs the best after post-processing in~\cite{baur2021autoencoders}, has difficulty reconstructing the normal images required for anomaly segmentation. By constrast, we obtain a mean Dice score of 63.67\% for the BraTS brain tumor segmentation and 32.24\% for the LiTS liver lesion segmentation, under the two-fold cross-validation settings for both datasets. In addition, we improve the Dice score for the liver lesion segmentaiton to 50.23\% using a simple post-processing scheme of open and closed sets.



\vspace{0.05in}
\noindent 
Overall, the contributions of our proposed method are summarized below:
\begin{itemize}[noitemsep,topsep=0pt]
\itemsep0em 
    \item Proposing an adversarial based framework for unsupervised anomaly segmentation, which bypasses the normal image reconstruction and works on anomaly detection directly. This framework presents a general clustering strategy to generate two selective cuts based on a reference image set with human knowledge. 
    \item To the best of our knowledge, our work is the first one to apply an unsupervised segmentation algorithm to the BraTS 2019 and LiTS liver lesion public datasets. Besides, our method outperforms the AnoGAN family and other popular methods presented in~\cite{baur2021autoencoders} on the publicly available MS-SEG2015 dataset.
\end{itemize}

\section{Adversarial-based Selective Cutting Network (ASC-Net)}

\subsection{Network Framework}
\label{sec:framework}

Figure~\ref{fig:overview} describes the framework of our proposed ASC-Net, which includes two components, i.e., the main module $M$ and the discriminator $D$, and one simple clustering step $T$ based on thresholding. Overall, the main module includes normal and anomaly branches to semantically reconstruct the original image for clustering, while the discriminator brings user-defined knowledge into the normal branch in the main module.

\begin{figure*}[t]
\centering
  \includegraphics[height=0.115\textwidth]{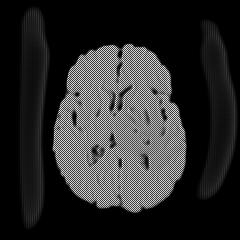}
  \includegraphics[height=0.115\textwidth,width=0.2\textwidth]{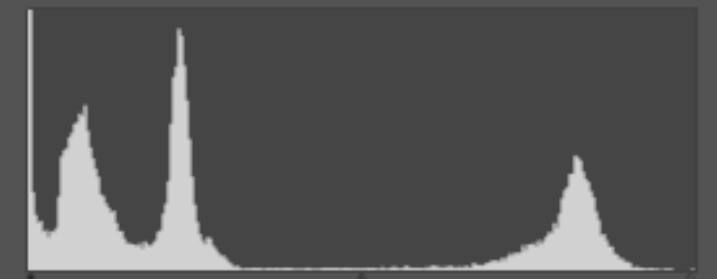}
  \includegraphics[height=0.115\textwidth]{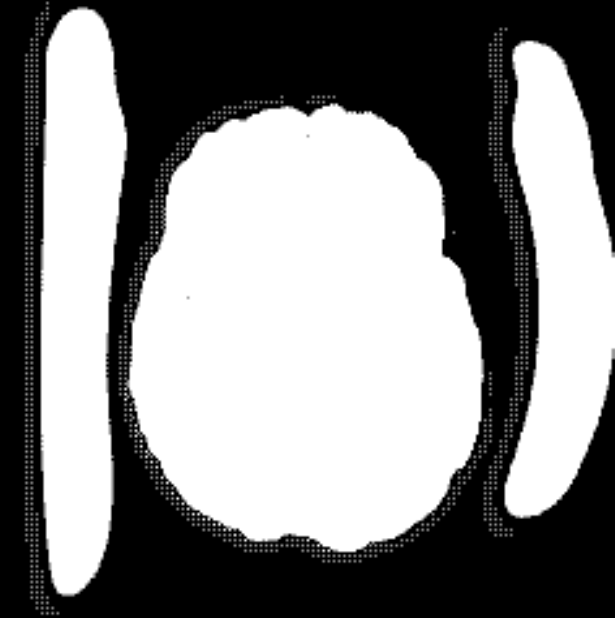}
  \includegraphics[height=0.115\textwidth]{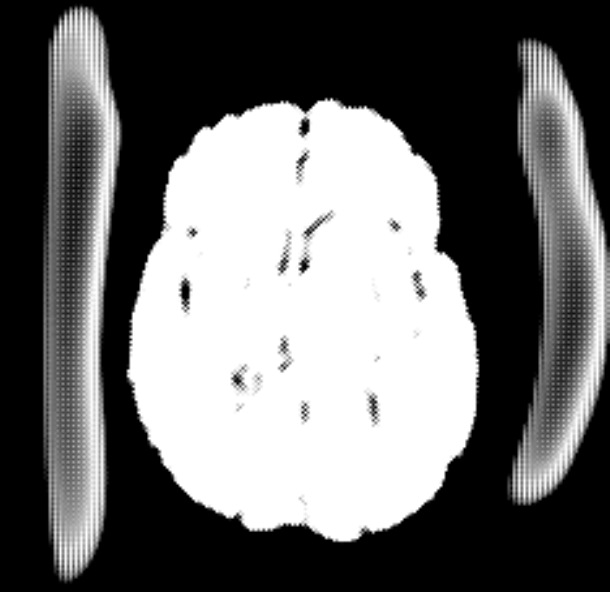}
  \includegraphics[height=0.115\textwidth]{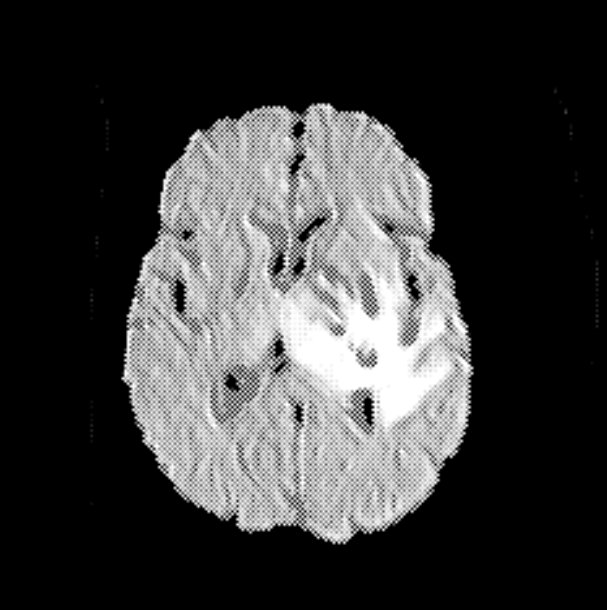}
  \includegraphics[height=0.115\textwidth]{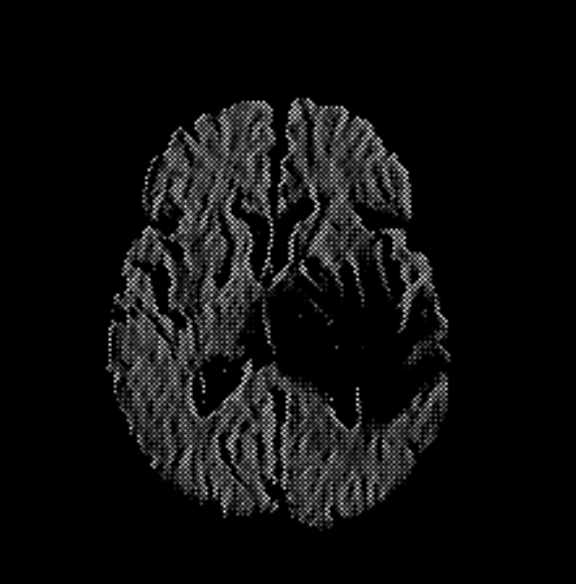}\\
  \includegraphics[height=0.116\textwidth]{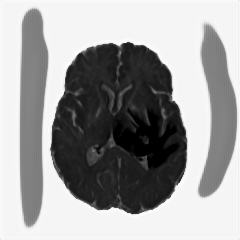}
  \includegraphics[height=0.116\textwidth,width=0.2\textwidth]{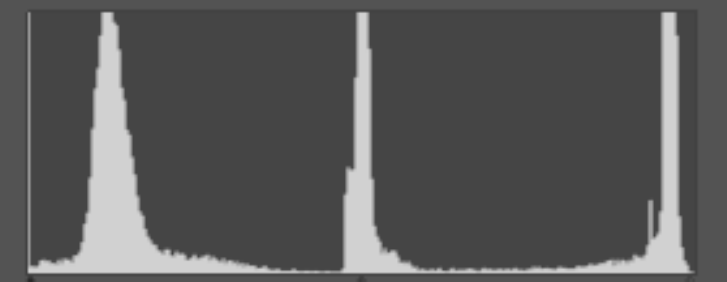}
  \includegraphics[height=0.116\textwidth]{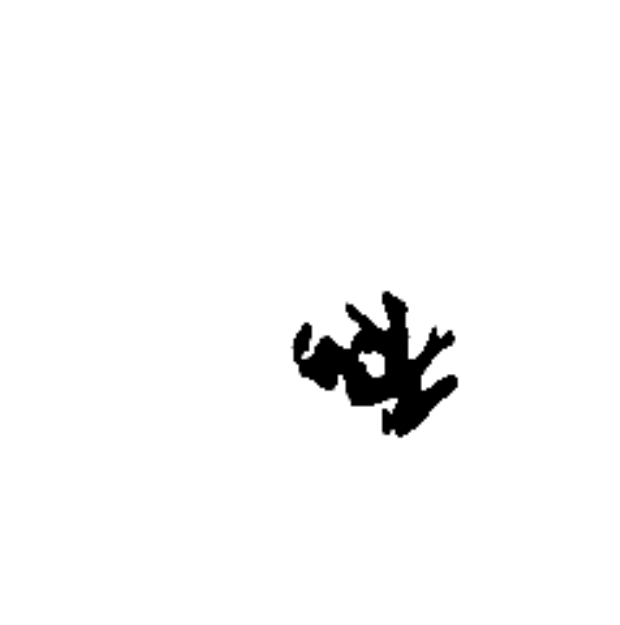}
  \includegraphics[height=0.116\textwidth]{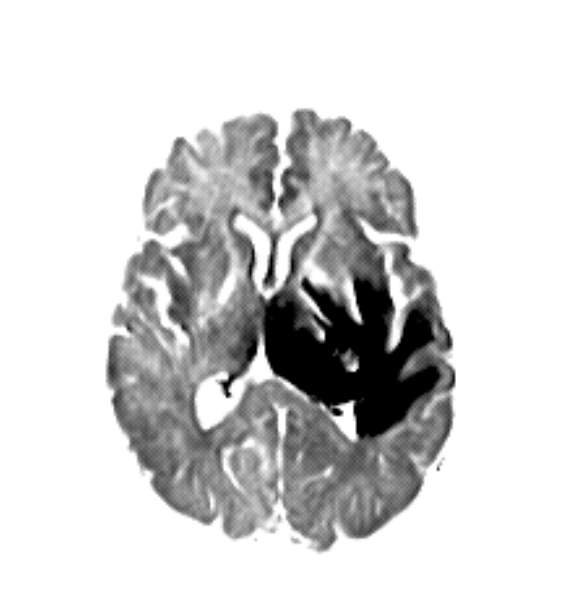}
  \includegraphics[height=0.116\textwidth]{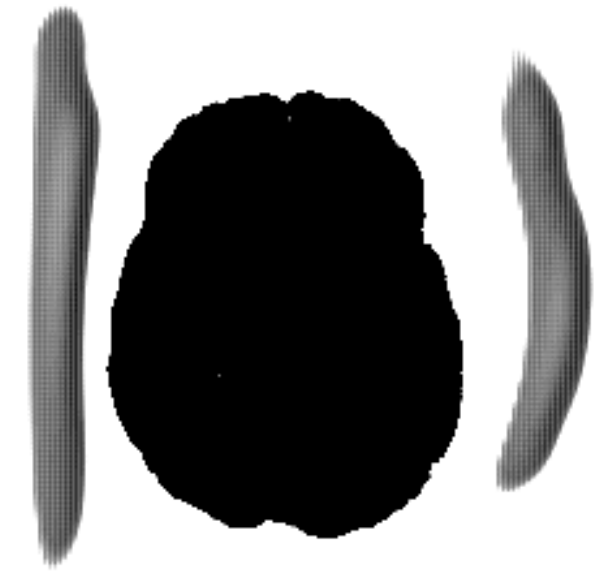}
  \includegraphics[height=0.116\textwidth]{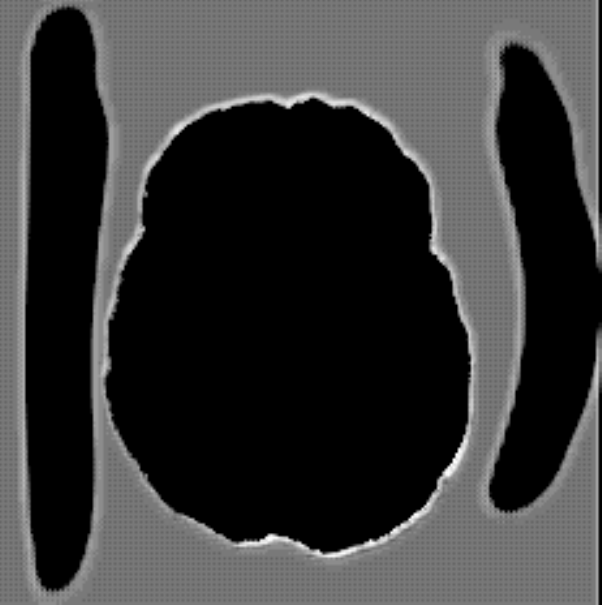}
\caption{Visualization of the “disjoincy” between images $I_{fc}$ (top) and $I_{wc}$ (bottom) generated by two cuts of ASC-Net. From left to right: the generated image, its histogram, and the following four columns representing the histogram equalized images of the thresholded peaks with the first peak being the first image, etc. The first peak of $I_{fc}$ is disjoint with the last peak of $I_{wc}$, etc.}
\label{fig:proof_dis}
\end{figure*}

\vspace{0.05in}
\noindent
\textbf{Main Module $M$.} The main module aims to generate two selective cuts, which guide a follow-up simple reconstruction of an input image to cluster image pixels based on intensity thresholding. The $M$ follows an encoder-decoder architecture like the U-Net, including one encoder and two decoders. The encoder $E$ extracts features of an input image $I_{in}$, which could be an image located within or outside of the reference distribution $\{I_{rd}\}$, a collection of normal images. 
One decoder in green (the second branch) is designed to generate a ``fence'' cut $C_f$ that is defined by an image fence formed by $\{I_{rd}\}$. The $C_f$ aims to generate an image $I_{fc}$ and tries to fool the discriminator $D$. The other decoder in blue (the first branch) is designed to generate another ``wild'' cut $C_w$, which captures leftover image content that is not included in $I_{fc}$. As a result, the $C_w$ produces another images $I_{wc}$ to complement the fence-cut output $I_{fc}$. The complementary relation between these two cuts $C_f$ and $C_w$ is enforced by a positive Dice loss discussed later. Figure~\ref{fig:proof_dis} demonstrates the ``disjoincy" of $I_{fc}$ and $I_{wc}$, like their complementary histogram distribution and different thresholded images at different peaks.



\begin{wrapfigure}{r}{0.49\textwidth}
\centering
  \includegraphics[height=0.115\textwidth,width=0.115\textwidth]{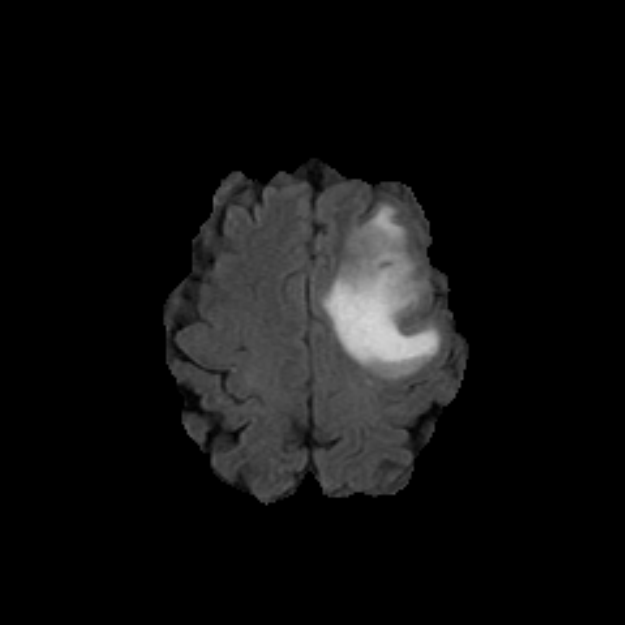}
  \includegraphics[height=0.115\textwidth,width=0.115\textwidth]{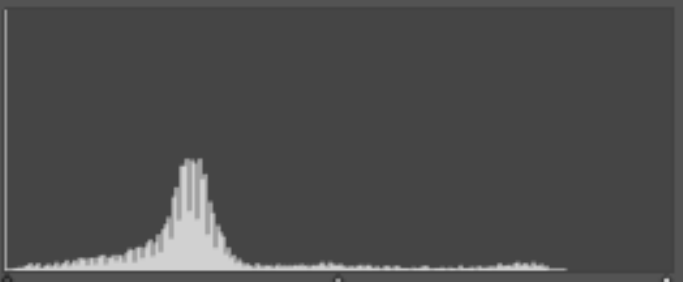}
  \includegraphics[height=0.115\textwidth,width=0.115\textwidth]{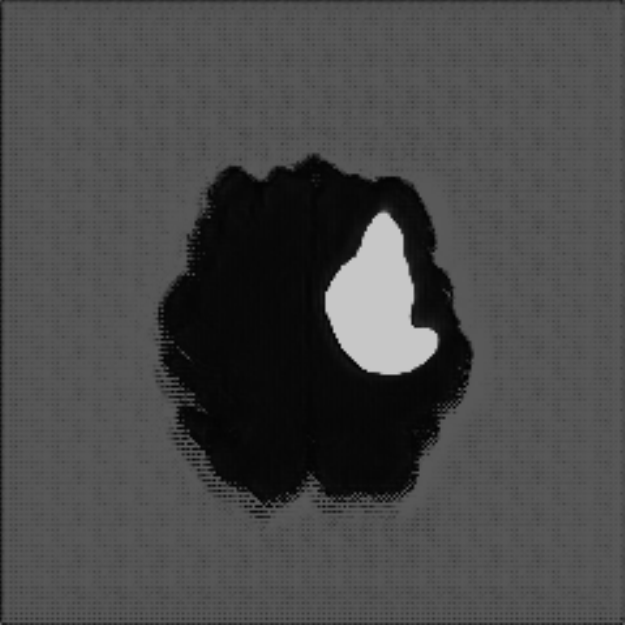}
  \includegraphics[height=0.115\textwidth,width=0.115\textwidth]{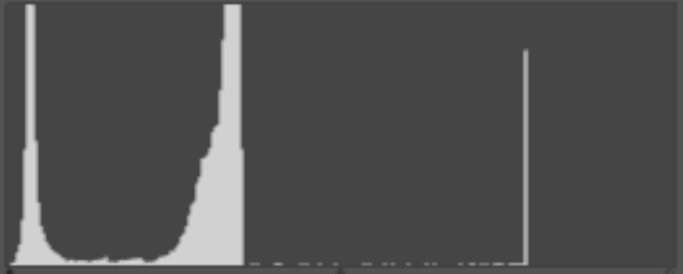}\\
    \includegraphics[height=0.115\textwidth,width=0.115\textwidth]{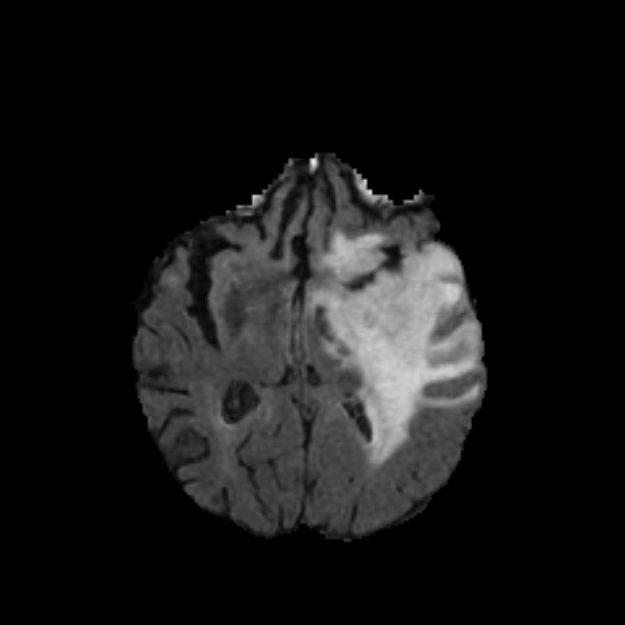}
  \includegraphics[height=0.115\textwidth,width=0.115\textwidth]{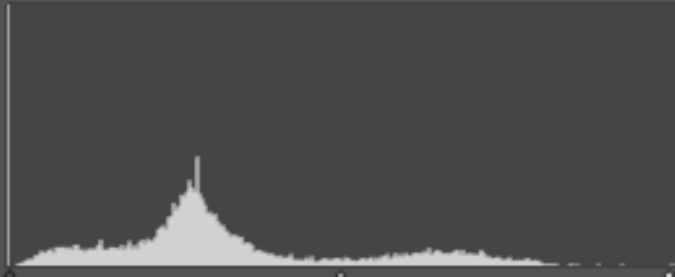}
  \includegraphics[height=0.115\textwidth,width=0.115\textwidth]{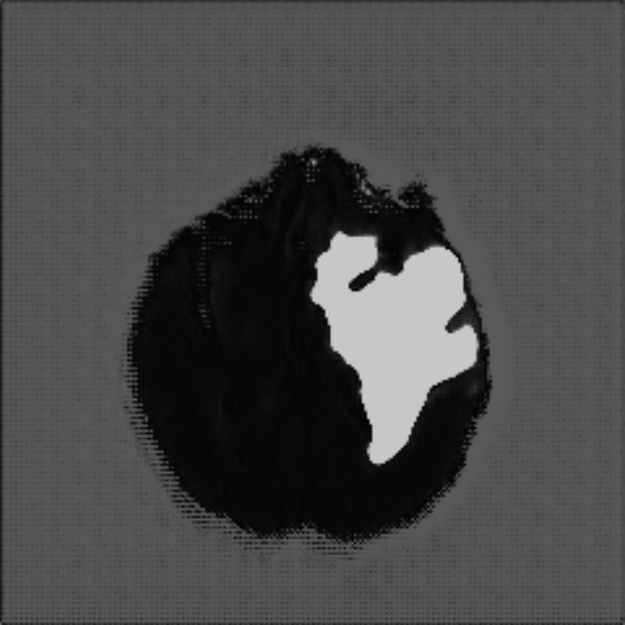}
  \includegraphics[height=0.115\textwidth,width=0.115\textwidth]{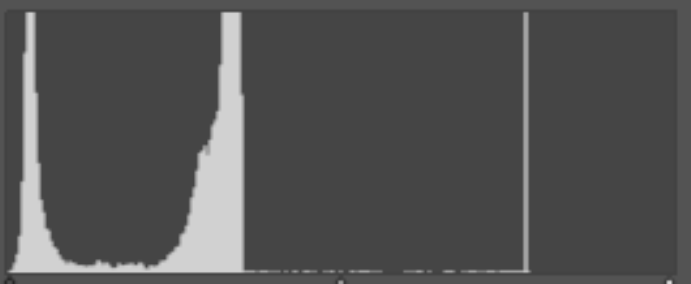}
\caption{Histogram comparison of two sample images. From left to right: the input image, its histogram, its reconstructed image using ASC-Net, and the histogram of the reconstructed image. The histograms of the input images vary greatly, while the ones of their reconstructions show peaks at similar ranges, which enables a thresholding based pixel-level separation.}
\label{fig:input_vs_op}
\end{wrapfigure}

The reconstructor $R$ consists of a $1 \times 1$ convolution layer with the Sigmoid as the activation function, which is applied on the concatenation of the two-cut outputs $I_{fc}$ and $I_{wc}$ to regenerate the input image $I_{in}$ back. This reconstructor $R$ ensures that the $C_f$ does not generate an image $I_{fc}$ far from the input image $I_{in}$ and also ensures that the $C_w$ does not generate an empty image $I_{wc}$ if the anomaly or novelty exists. Figure~\ref{fig:input_vs_op} shows the histogram separation of the reconstructed images, compared to the original input images which present complex histogram peaks and have difficulty in separating the brain tumor from backgorund and other tissues via a simple thresholding. The discontinuous histogram distribution of $I_{ro}$ is inherited from the two generated sub-images $I_{fc}$ and $I_{wc}$ through a simple weighted combination. As a result, the segmentation task becomes relatively easy to be done on the reconstructed image $I_{ro}$.

\vspace{0.05in}
\noindent
\textbf{Discriminator $D$.} The GAN discriminator tries to distinguish the generated image $I_{fc}$, according to a reference distribution $R_d$ defined by a set of images $\{I_{rd}\}$, which are provided by the user or experts. The $R_d$ typically includes images collected from the same group, for instance, normal brain scans, which share similar structures and lie on a manifold. Introducing $D$ allows us to incorporate our vague prior knowledge about a task into a deep neural network. Typically, it is non-trivial to explicitly formulate such prior knowledge; however, it could be implicitly represented by a selected image set. The $R_d$ is an essential component that makes our ASC-Net possible to generate selective cuts according to the user's input, without requiring other supervisions. 

\vspace{0.05in}
\noindent
\textbf{Thresholding $T$.}
To cluster the reconstructed image $I_{ro}$ into two groups at the pixel level, we choose the thresholding approach with the threshold values obtained using the histogram of $I_{ro}$. We observed that for an anomaly that is often brighter than the surrounding tissues like the BraTS brain tumor, the intensity value at the rightmost peak of the histogram is a desired threshold; while an opposite case like darker LiTS liver lesions, the value at the leftmost peak would be the threshold. We also observed that the histograms of the reconstructed images for different inputs reflect the same cut-off point for the left or right peaks, which allows using one threshold for an entire dataset.

\vspace{0.05in}
\noindent
\textbf{Loss Functions.}
The main module $M$ includes three loss functions: (i) the image generation loss for $C_f$ ($Loss_{C_f}$), (ii) the ``disjoincy" loss between $C_f$ and $C_w$ ($Loss_{C_w}$), and (iii) the reconstruction loss ($Loss_{R}$).   
In particular, the $C_f$ tries to generate an image $I_{fc}$ that fools the discriminator $D$ by minimizing $Loss_{C_f} = \frac{1}{n}\sum_{i=1}^{n} |D(I_{fc}^{(i)})-1|$. Here, $n$ is the number of samples in the training batch. The $C_w$ tries to generate an image $I_{wc}$ that is complement to $I_{fc}$ by minimizing the positive Dice score $Loss_{C_w} = \frac{2 |I_{fc} \cap I_{wc}|}{|I_{fc}| + |I_{wc}|}$.
The last reconstruction takes an Mean-Squared-Error (MSE) loss between the input image $I_{in}$ and the reconstructed image $I_{ro}$:
    $Loss_R = \frac{1}{n}\sum_{i=1}^{n}\|I_{in}^{(i)} - I_{ro}^{(i)}\|_2^{2}$. 
The discriminator $D$ tries to reject the the $C_f$ output $I_{fc}$ but accept the images from the reference distribution $R_d$, by minimizing the following loss function:
    $Loss_D = \frac{1}{n+m} \left(\sum_{i=1}^{n} |D(I_{fc}^{(i)})-(-1)| +\sum_{i=1}^{m} |D(I_{R_d}^{(i)})-1| \right)$.
Here, $m$ is the number of the images in $R_d$. Even though $D$ and $C_{fc}$ are tied in an adversarial setup, here we do not use the Earth Mover distance~\cite{rubner2000earth} in the loss function, since we would like $D$ to identify both positive samples and negative samples with equal precision. Therefore, we use Mean Absolute Error (MAE) instead. 


\subsection{Architecture Details and Training Scheme}
\label{sec:training}
We use the same network architecture for all of our experiments as shown in Fig.~\ref{fig:overview}. 
The encoder $E$ consists of four blocks of two convolution layers with a filter size of $(3,3)$ followed by a max pooling layer with a filter size of $(2,2)$ and batch normalization after every convolution layer. After every pooling layer we also introduce a dropout of 0.3. The number of feature maps in each of the convolution layer of a block are 32, 64, 128, and 256. Following these blocks is a transition layer of two convolution layers with feature maps of size 512 followed by batch normalization layers. 
The $C_{fc}$ and $C_{wc}$ decoders are connected to the $E$ and mirror the layers with the pooling layers replaced with 2D transposed convolutional layers, which have the same number of feature maps as the blocks mirror those in the encoder. Similar to a U-Net, we also introduce skip connections across similar levels in the encoder and decoders. The reconstructor $R$ is simply a Sigmoid layer applied to the concatenation of $I_{fc}$ and $I_{wc}$, resulting in a simplified CompNet~\cite{dey2018compnet}. 
The Discriminator $D$ mimics the architecture of the $E$, except for the last layer where a dense layer is used for classification. All the intermediate layers have ReLU activation function and the final output layers have the Sigmoid activation. The only exception is the output of the discriminator $D$, which has a Tanh activation function to separate $I_{fc}$ and images from the $R_d$ to the maximum extent. 

\vspace{0.05in}
\noindent
We use Keras with Tensorflow backend and Adam optimizer with a learning rate of 5e-5 to implement our architecture. 
We follow two distinct training stages:
\begin{itemize}[noitemsep, nolistsep]
\item In the {\it first stage}, we train $D$ and $M$ in cycles. We start training $D$ with $\{R_d\}$ with True labels and $\{I_{fc}\}$ with False labels. These training samples are shuffled randomly. Following $D$, we train $M$ with $\{I_{in}\}$ as input and the weights of $D$ frozen while preserving the connection between $\{I_{fc}\}$ and $D$. The objective of the $M$ is to morph the appearance of $\{I_{in}\}$ into $\{I_{fc}\}$ to fool $D$ with the frozen weights. We call these two steps one cycle, and in each step there may be more than one epochs of training for $M$ or $D$. 



\item In the {\it second stage}, $M$ and $D$ continue to be trained alternatively; however, the input images to $D$ are changed, since the training purpose at this stage is to focus on the differences between the $\{R_d\}$ and $\{I_{in}\}$, while ignoring the noisy biases created by the $M$ in transforming $\{I_{in}\}$ to $\{I_{fc}\}$. To achieve this, we augment the reference distribution $\{R_d\}$ with its generated images via $M$, i.e., $\{I_{fc}(R_d)\}$. We treat them as true images, and the union set $\{R_d \cup I_{fc}(R_d)\}$ is used to update $D$. 

\end{itemize}

\vspace{0.05in}
\noindent
\textbf{Runtime Analysis.} We use two Nvidia TitanX GPUs and on average a discriminator cycle takes 2.5 ms to process a single 2D image slice with size of 240 $\times$ 240, while the main module cycle takes 15.5 ms to process a single 2D image slice during training.
%
%
%
%
\section{Applications}
We evaluate our model on three unsupervised anomaly segmentation tasks: MS lesion segmentation, brain tumor segmentation, and liver lesion segmentation. We use the MS-SEG2015~\cite{carass2017longitudinal} training set,  BraTS~\cite{bakas2017advancing,bakas2018identifying,menze2014multimodal}, and  LiTS~\cite{bilic2019liver} datasets in these tasks. 


\vspace{0.05in}
\noindent
\textbf{MS-SEG2015.} The training set consists of 21 scans from 5 subjects with each scan dimensions of $181 \times217 \times 181$. We resize the axial slices to $160 \times 160$, so that we can share the same network design as the rest of the experiments. 

\vspace{0.05in}
\noindent
\textbf{BraTS 2019.} This dataset consists of 335 T1-w MRI brain scans collected from 259 subjects with high grade Glioma and 76 subjects with low grade Gliomas in the training set. The 3D dimensions of the images are $240 \times 240 \times 155$. 

\vspace{0.05in}
\noindent
\textbf{LiTS.} The training set of LiTS consists of 130 abdomen
CT scans of patients with liver lesions, collected from multiple institutions. Each scan has a varying number of slices with dimensions of 512$\times$512. We resize these CT slices to $240 \times 240$ to share the same network architecture with other tasks. 


\vspace{0.05in}
\noindent
For all experiments, the image intensity is normalized to $[0, 1]$ over the 3D volume; however, we perform the 3D segmentation task in the slice-by-slice manner using axial slices. To balance the sample size in $I_{in}$ and $R_d$, we randomly sample and duplicate the number difference to the respective set.

\vspace{0.05in}
\noindent
\textbf{MS Lesion Segmentation.} In this task, we randomly sample $2870$ non-tumor, non-zero, Brats-2019 training set slices to make our reference distribution $R_d$ as in~\cite{baur2021autoencoders}, while they use their own privately annotated healthy dataset. Meanwhile, the $2870$ non zero 2D slices of the MS-SEG2015 training set are used in the main module $M$. We train this network using three cycles in the first stage and one cycle in the second stage and take the threshold at 254 intensity based on the right most peak of the image histogram. 

\begin{figure*}[t]
\centering
\setlength\tabcolsep{1.5pt}
\begin{tabular}{ccccccc}
$I_{in}$ & $I_{fc}$ & $I_{wc}$ & $I_{ro}$ & $M_{gt}$& $M_{est}$ & $M_{est} \cap I_{in}$ \\

\includegraphics[width=0.13\textwidth]{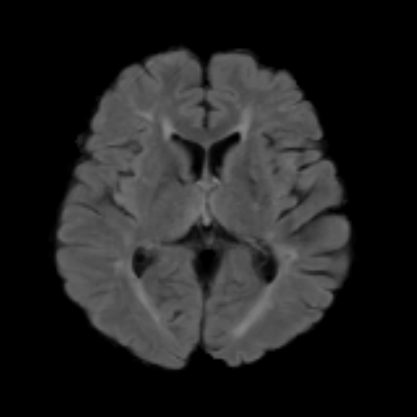} &
\includegraphics[width=0.13\textwidth]{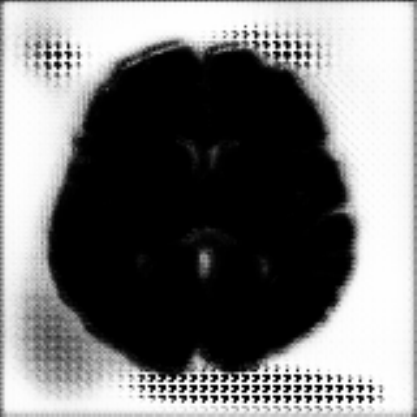} &
\includegraphics[width=0.13\textwidth]{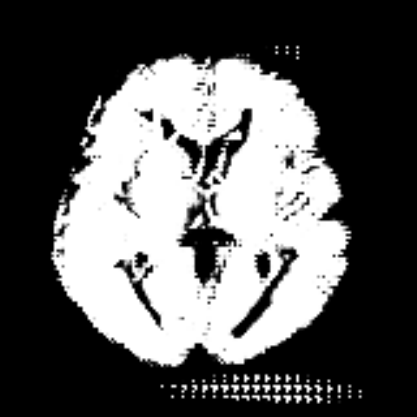} &
\includegraphics[width=0.13\textwidth]{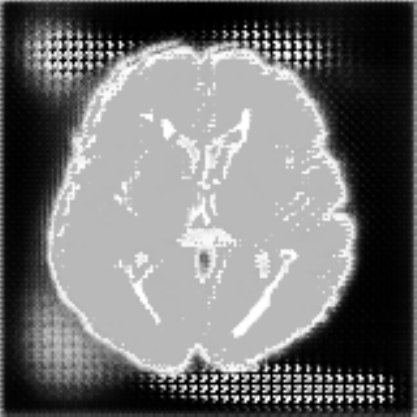} &
\includegraphics[width=0.13\textwidth]{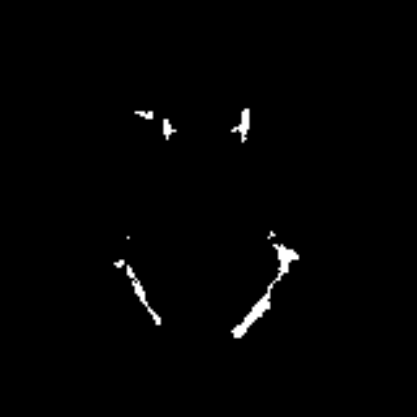} &
\includegraphics[width=0.13\textwidth]{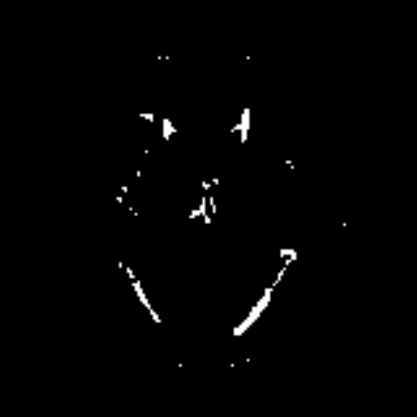} &
\includegraphics[width=0.13\textwidth]{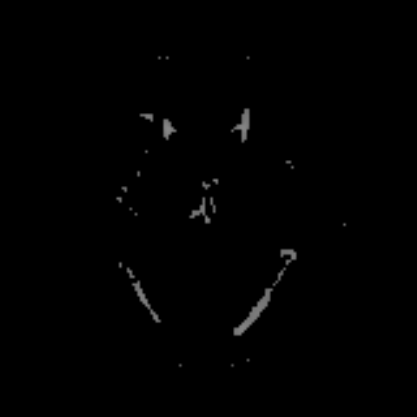} \\

\includegraphics[width=0.13\textwidth]{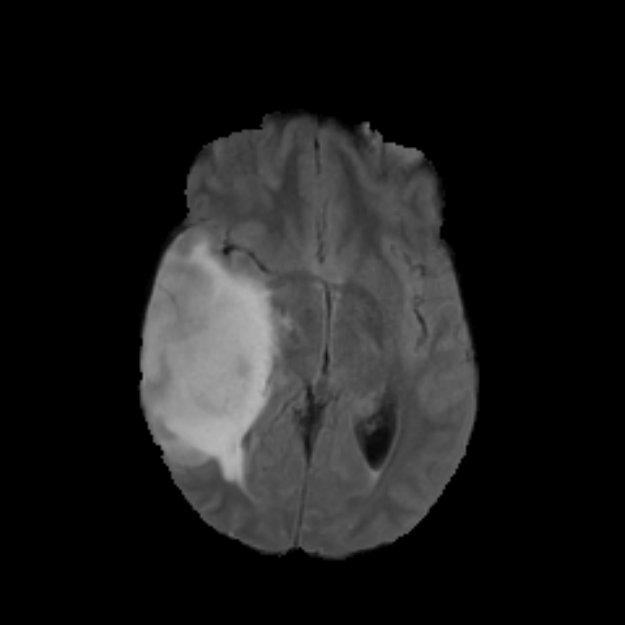} &
\includegraphics[width=0.13\textwidth]{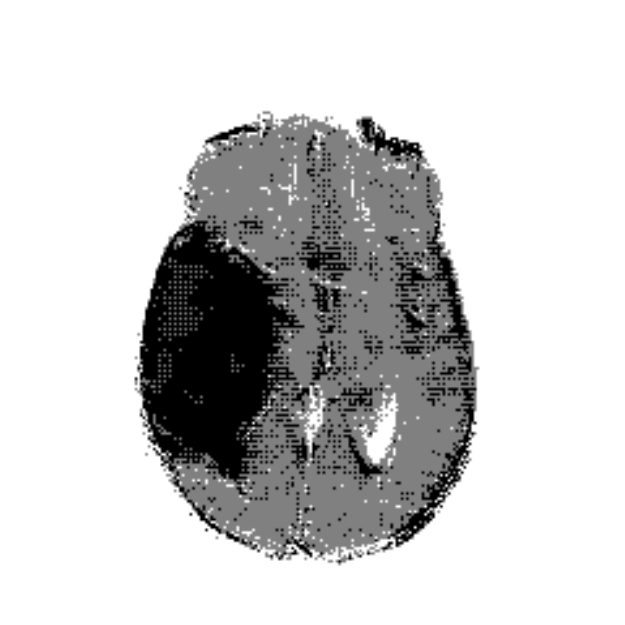} &
\includegraphics[width=0.13\textwidth]{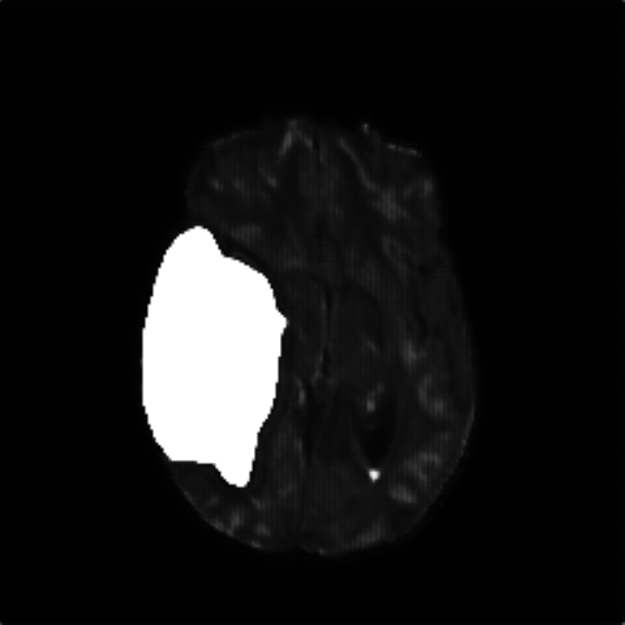} &
\includegraphics[width=0.13\textwidth]{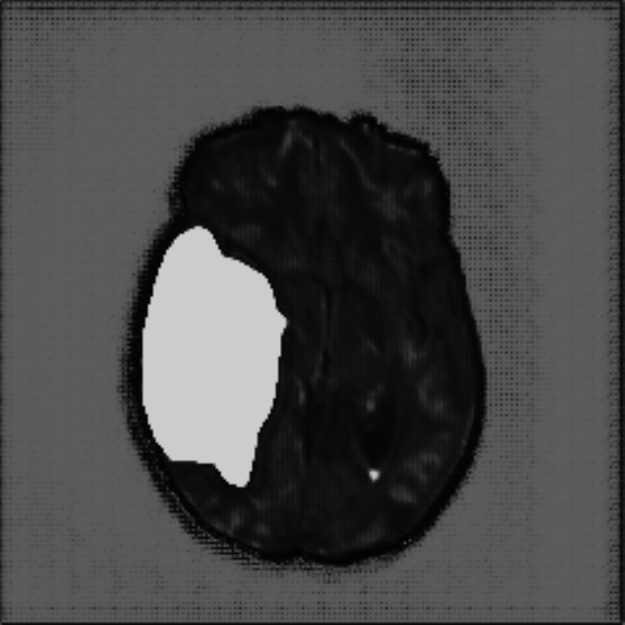} &
\includegraphics[width=0.13\textwidth]{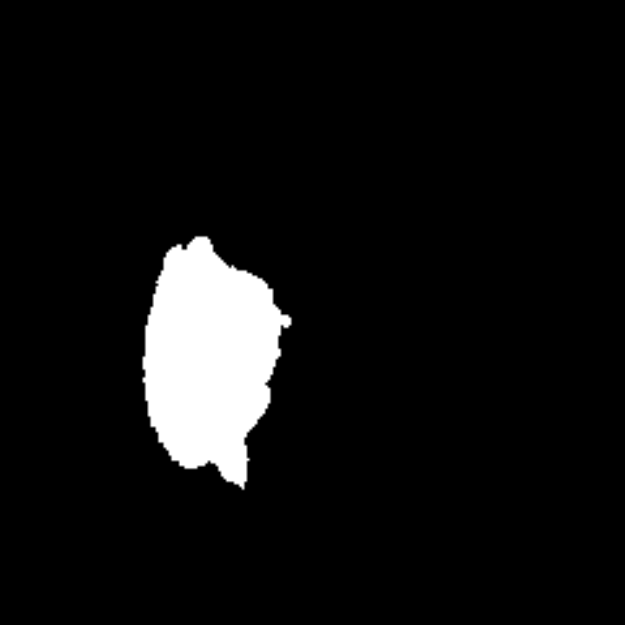} &
\includegraphics[width=0.13\textwidth]{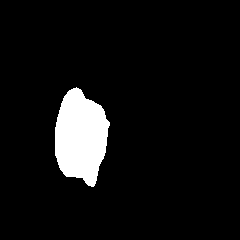} &
\includegraphics[width=0.13\textwidth]{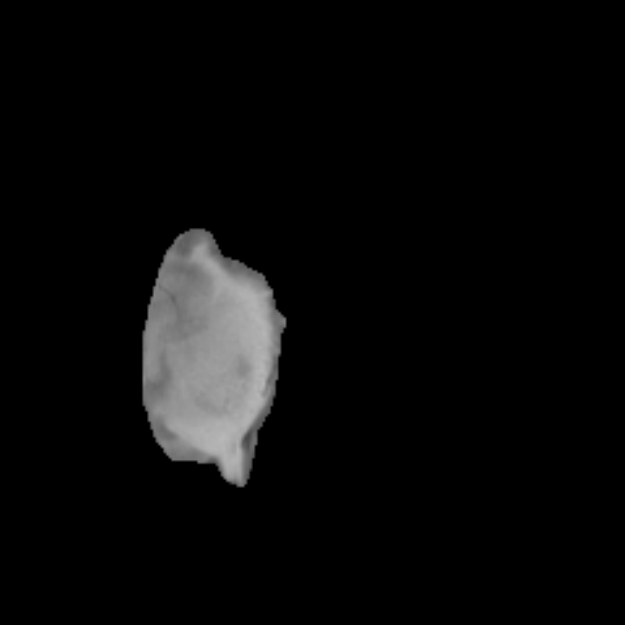} \\


\includegraphics[width=0.13\textwidth]{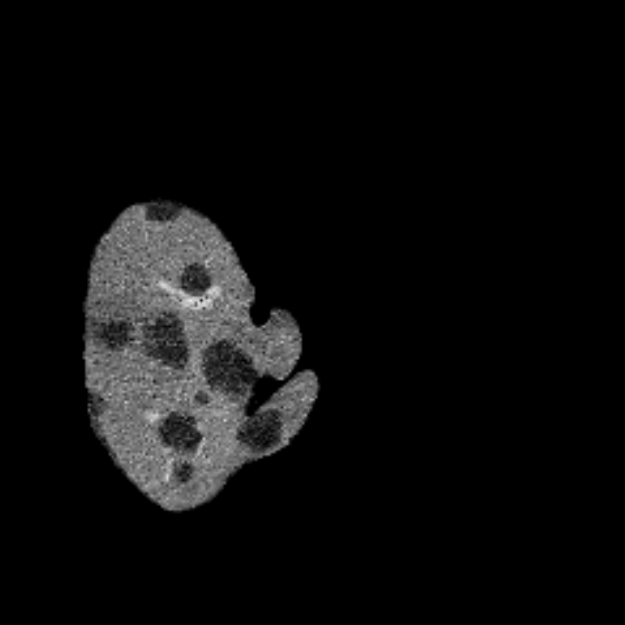} &
\includegraphics[width=0.13\textwidth]{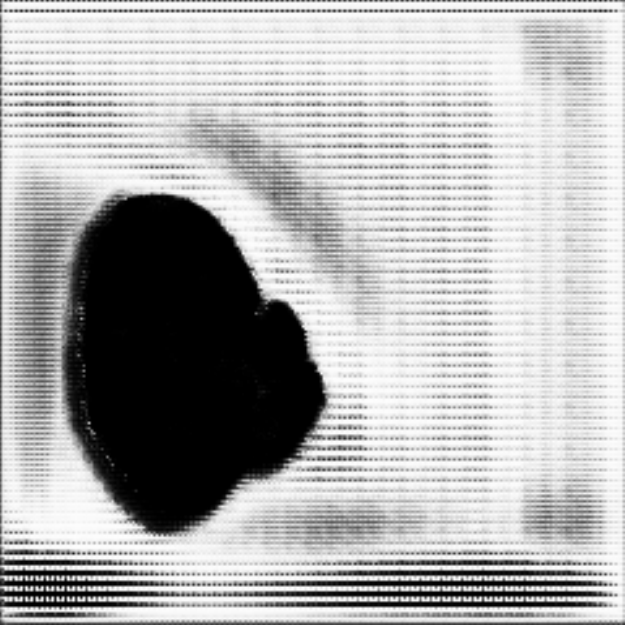} &
\includegraphics[width=0.13\textwidth]{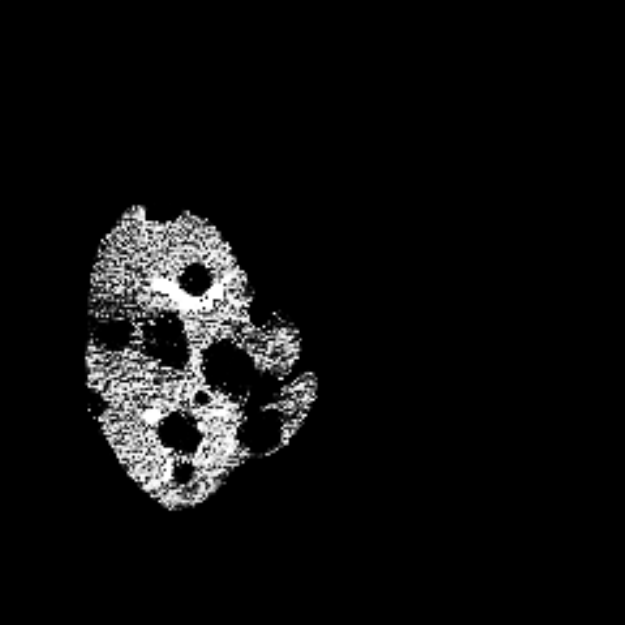} &
\includegraphics[width=0.13\textwidth]{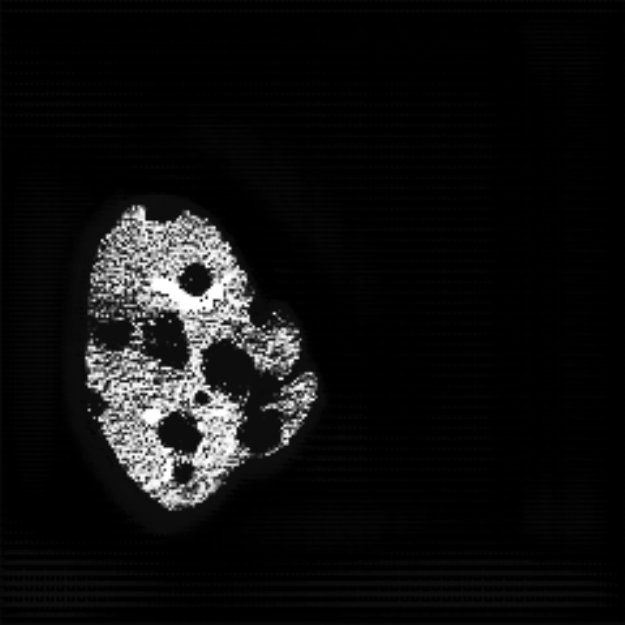} &
\includegraphics[width=0.13\textwidth]{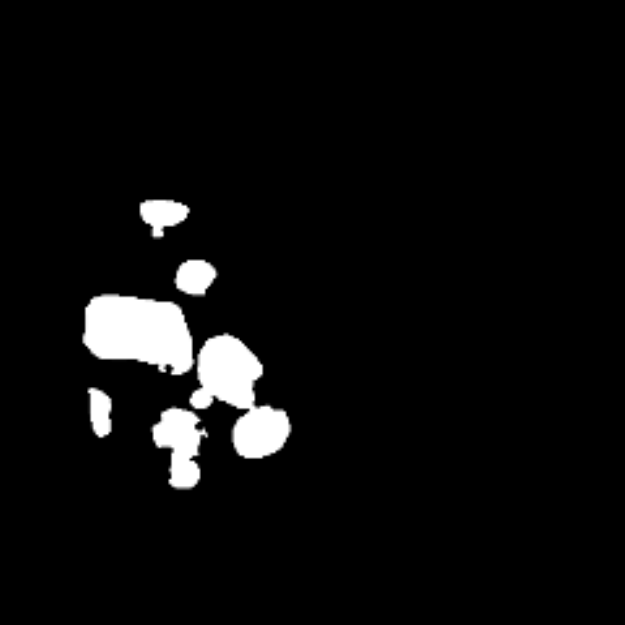} &
\includegraphics[width=0.13\textwidth]{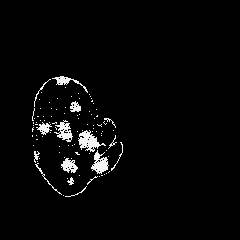} &
\includegraphics[width=0.13\textwidth]{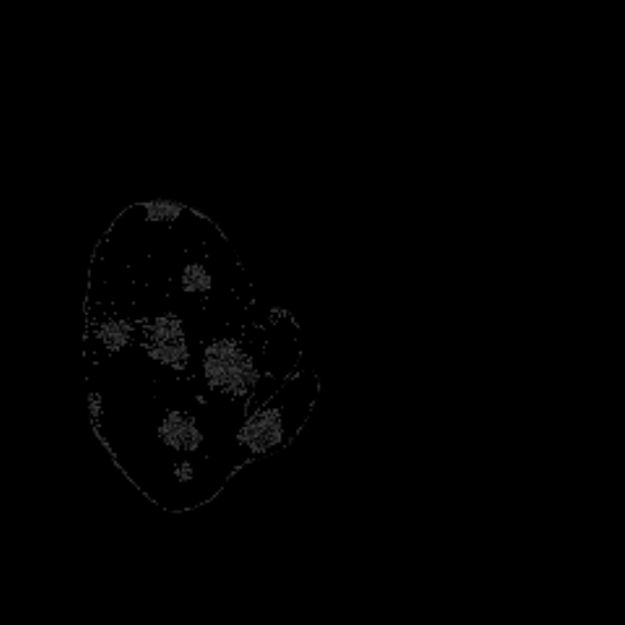} 
\end{tabular}
\caption{Sample results of MS-SEG2015, Brats-2019 and LiTS (top to bottom) obtained from the various branches of the network. The $I_{fc}$ in the second row is contrast enhanced to present the content contained in the brain region. None of these include any of the post processed images.} 
\label{fig:result_brats}
\end{figure*}

We obtain an average Dice score of 32.94\% without any post processing. By using a simple post-processing with erosion and dilation\footnote{We use this operator to improve the connectivity of the generated anomaly mask.} with $5\times5$ filters, this number improves to 48.20\% Dice score. In comparison, a similar study conducted by~\cite{baur2021autoencoders} consisting of a multitude of algorithms including AnoVAEGAN \cite{baur2018deep} and f-AnoGANS, obtained a best mean score of 27.8\% Dice after post processing by f-AnoGANS. Before post processing the best method was Constrained AutoEncoder~\cite{chen2018unsupervised} with a score of 9.7\% Dice. 
Sample images of our method are included in Fig.~\ref{fig:result_brats}


    


\vspace{0.05in}
\noindent
\textbf{Brain Tumor Segmentation.} In this task, we perform patient-wise two-fold cross-validation on the Brats-2019 training set. In each training fold, we use a 90/10 split after removing empty slices. The 2D slices from the 90\% split without tumors are used to make our reference distribution $R_d$; while the 2D slices with tumors from the 90\% split and all the slices from the 10\% split are used for training our model. As a result, the sample size of $R_d$ for fold one and two amounts to 11,745 and 12,407 respectively, while the size of $I_{in}$ amounts to 11,364 and 10,786, respectively. We train this network using two cycles in the first stage and one cycle in the second stage.






\begin{figure*}[t]
\centering
\includegraphics[width=0.96\textwidth]{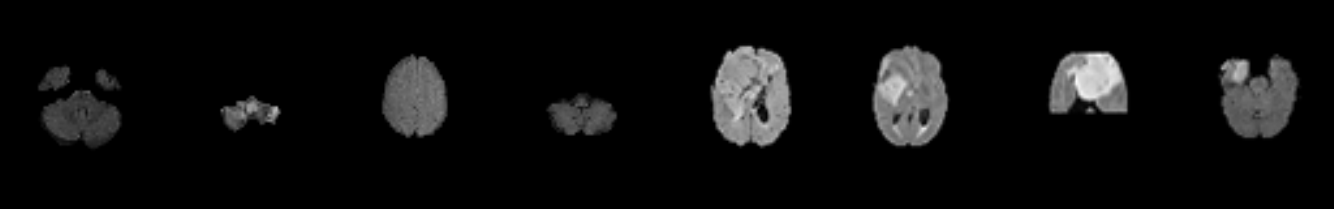}
\includegraphics[width=0.96\textwidth]{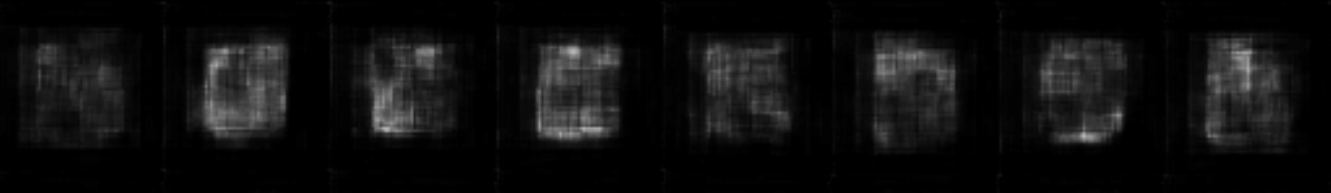}
\caption{Query images (top) and their reconstructions (bottom) using f-AnoGANs~\cite{schlegl2019f}. }
\label{fig:anogans_fail}
\end{figure*}

We obtain an average Dice score of 63.67\% for the brain tumor segmentation. 
Figure~\ref{fig:result_brats} shows samples generated by our ASC-Net. Figure~\ref{fig:anogans_fail} shows our attempt to apply f-AnoGANs~\cite{schlegl2019f} by following their online instructions. The failure of AnoGANs in the reconstruction brings to light the issue with the regeneration based methods and the complexity and stability of GAN based image reconstruction. 


\vspace{0.05in}
\noindent
\textbf{Liver Lesion Segmentation.} To generate the image data for this task, we remove the non-liver region by using the liver mask generated by CompNet~\cite{dey2018compnet} and take all non-zero images. We have 11,926 2D slices without liver lesions used in the reference distribution $R_d$. The remaining 6,991 images are then used for training the model. We perform slice-by-slice two-fold cross-validation and train the network using two cycles in both first and second stages. To extract the liver lesions, we first mask out the noises in the non-liver region of the reconstructed image $I_{ro}$ and then invert the image to take a threshold value at 242, the rightmost peak of the inverted image.

We obtain an average Dice score of 32.24\% for this liver lesion segmentation, 
which improves to 50.23\% by using a simple post processing scheme of erosion and dilation with $5\times5$ filter. Sampled results are shown in Fig.~\ref{fig:result_brats}. In comparison, a recent study~\cite{dey2020hybrid} reports a cross-validation result of 67.3\% under a supervised setting.  
Note that the annotation in the LiTS lesion dataset is imperfect with missing small lesions~\cite{chlebus2018automatic,dey2020hybrid}. Since we use the imperfect annotation to select images for the reference distribution, some slices with small lesions may be included and treated as normal examples.

\begin{figure}[t]
\centering
  \includegraphics[width=0.105\textwidth]{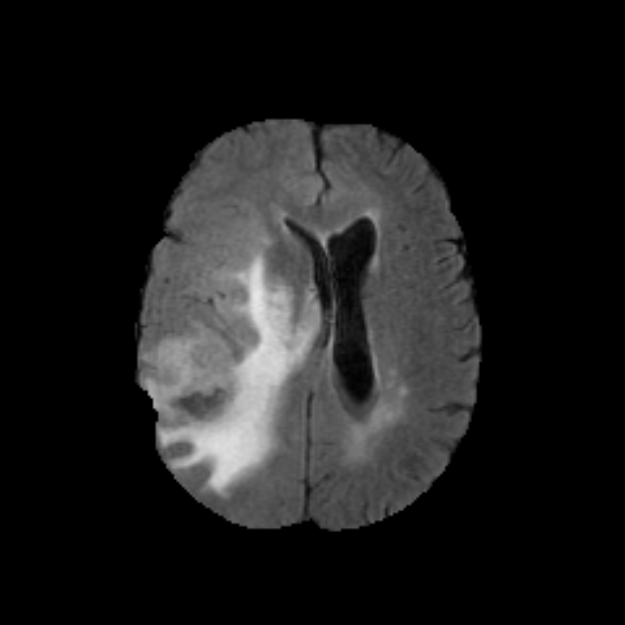}
  \includegraphics[width=0.105\textwidth]{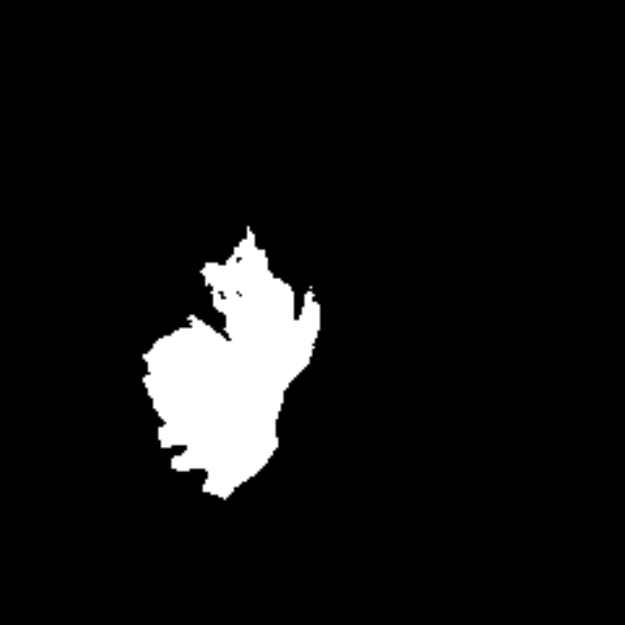}
  \includegraphics[width=0.105\textwidth]{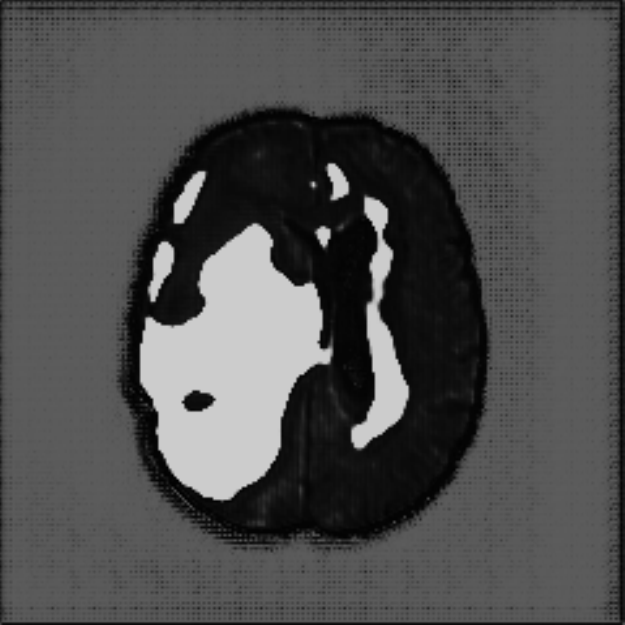}
    \includegraphics[width=0.105\textwidth]{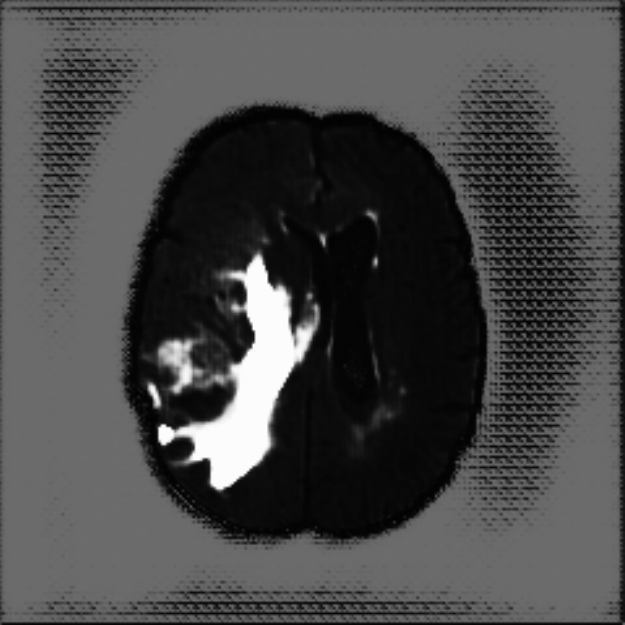}
  \includegraphics[width=0.105\textwidth]{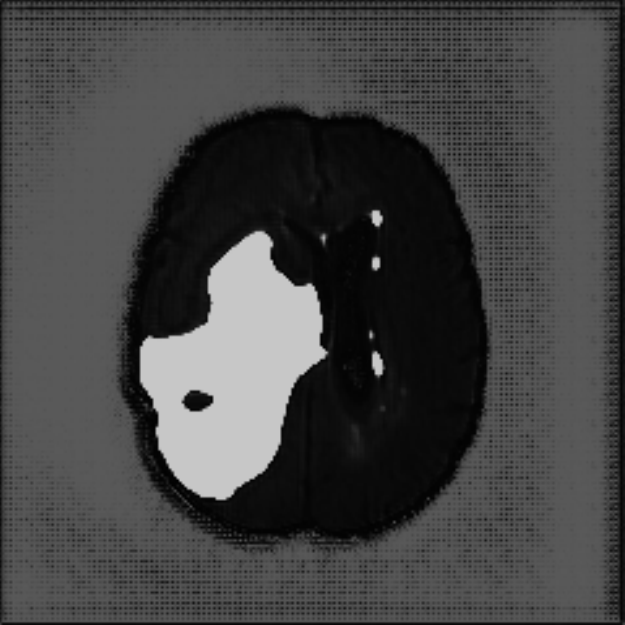}
  \includegraphics[width=0.105\textwidth]{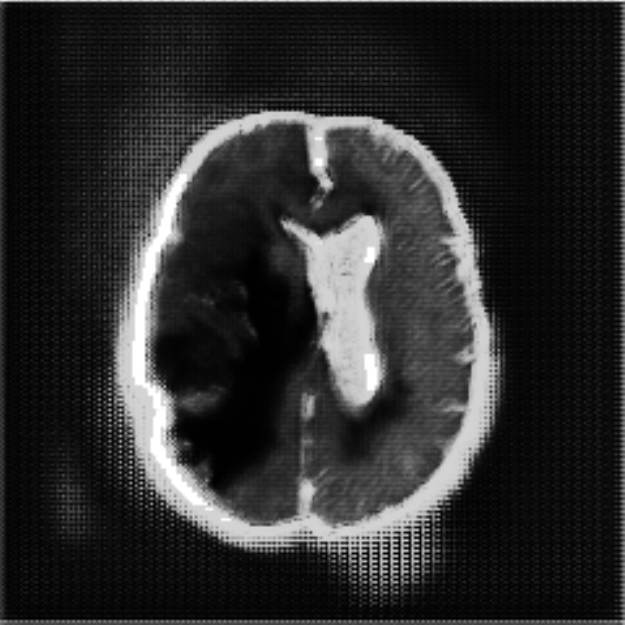}
    \includegraphics[width=0.105\textwidth]{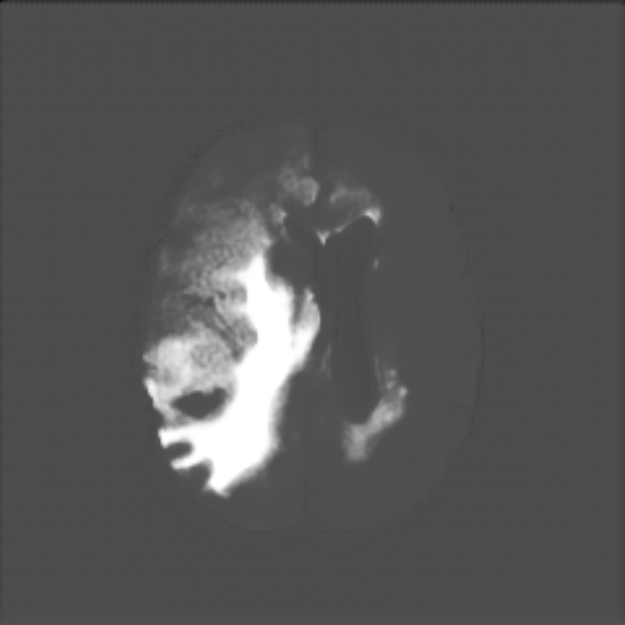}
  \includegraphics[width=0.105\textwidth]{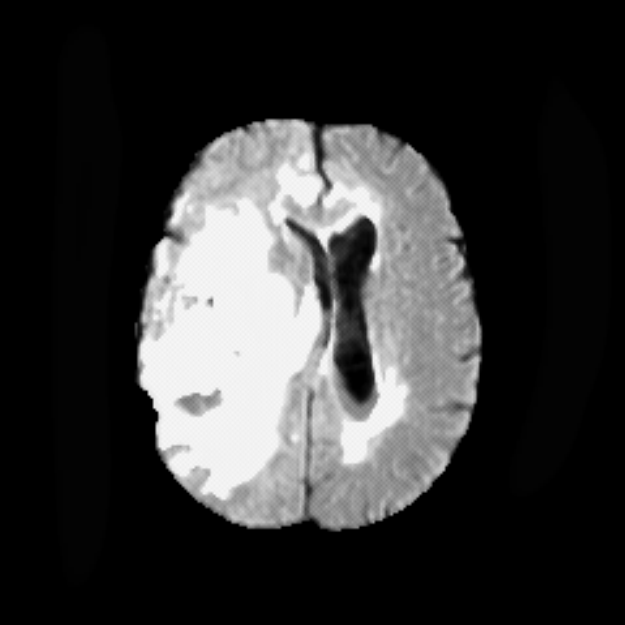}
  \includegraphics[width=0.105\textwidth]{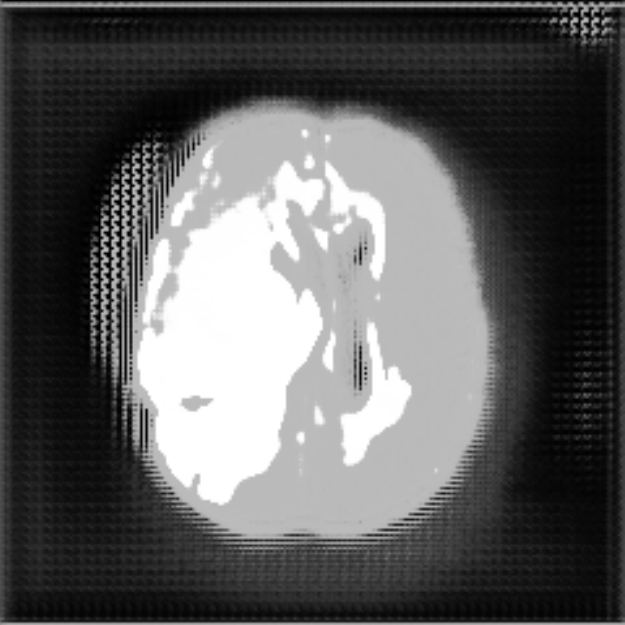}
\caption{\textbf{Stability}: The first image is the input image, the second is the ground truth. The rest of images are reconstruction from various re-runs of the framework with variable training cycles and stage. All runs are able to isolate the anomaly in question.}
\label{fig:stability}
\end{figure}

\section{Discussion and Future Work}

In this paper we have presented a framework that performs two-cut split in an unsupervised fashion guided by an reference distribution using GANs. Unlike the methods in the AnoGAN family which operate as a reconstruction-based method and needs faithful reconstruction of normal images to function properly, we treat the anomaly segmentation as a constrained two-cut problem that requires a semantical and reduced reconstruction for clustering. Our ASC-Net focuses on the anomaly detection with the normal image reconstruction as a byproduct, thus still producing competitive results where reconstruction dependent methods such as f-AnoGAN fails to work on. The current version of our ASC-Net aims to solve the two-cut problem, which will be tasked to handle more than two selective cuts in the future. Theoretical understanding of the proposed network is also required, which is left as a future work.

\vspace{0.05in}
\noindent
\textbf{Limitations and Opportunities.} One reason of our low Dice scores could be that we had to select non-tumor or normal slices as our reference distribution, which does not account for other co-morbidities. This affects the performance of the framework as it has no other guidance and would consider co-morbidities as an anomaly as well. However, this provides possibility of bringing other anomalies into the users' attention.

\begin{wrapfigure}{r}{0.35\textwidth}
\centering
  \includegraphics[width=0.35\textwidth]{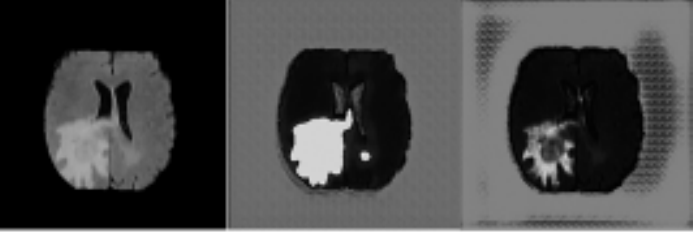}\\
  \includegraphics[width=0.35\textwidth]{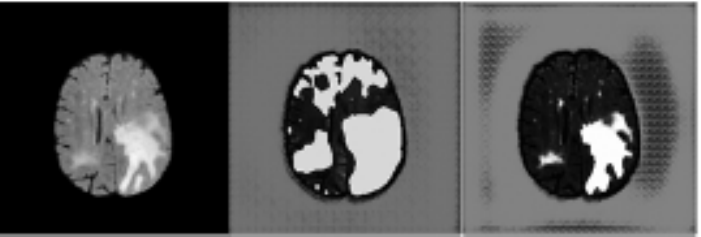}
\caption{Termination of network training affects the reconstruction result. Left to right columns in each row: the input image, the image reconstructed via two cycles in the first stage and one in the second stage, and the image reconstructed via adding one cycle in the second stage.}
\label{fig:train_more}
\end{wrapfigure}

\vspace{0.05in}
\noindent
\textbf{Termination and Stability.} The termination point of this network training is periodic. The general guideline is that the peaks should be well separated and we terminate our algorithm at three or four peak separation. However, continuing to train further may not always result in the improvement for the purpose of segmentation due to accumulation of holes as shown in  Fig.~\ref{fig:train_more}, even though visually the anomaly is captured in more intricate detail. We however encourage training longer as it reduces false positive and provide detailed anomaly reconstruction, though the Dice metric might not account for it. In our experiments, we specify the number of cycles in each stage. However, due to the random nature of the algorithm and the lack of a particular purpose and guidance, the peak separation may occur much earlier, then training should be stopped accordingly. The reported network in our Brats-2019 experiments has an average Dice score of $6\%$ over the network trained longer as shown in Fig.~\ref{fig:train_more}. Regarding the stability, Figure~\ref{fig:stability} demonstrates an anomaly estimated by different networks that are trained with different number of training cycles. We observe that while the appearance of $I_{ro}$ changes, we still obtain the anomaly as a separate cut since our framework works without depending on the quality of reconstruction.

\subsection*{Acknowledgements} This work was supported by NSF 1755970 and Shanghai Municipal Science and Technology Major Project 2021SHZDZX0102.

{\small
\bibliographystyle{plain}
\bibliography{refs}

\begin{thebibliography}{10}

\bibitem{bakas2017advancing}
Spyridon Bakas, Hamed Akbari, Aristeidis Sotiras, Michel Bilello, Martin
  Rozycki, Justin~S Kirby, John~B Freymann, Keyvan Farahani, and Christos
  Davatzikos.
\newblock Advancing the cancer genome atlas glioma mri collections with expert
  segmentation labels and radiomic features.
\newblock {\em Scientific data}, 4:170117, 2017.

\bibitem{bakas2018identifying}
Spyridon Bakas, Mauricio Reyes, Andras Jakab, Stefan Bauer, Markus Rempfler,
  Alessandro Crimi, Russell~Takeshi Shinohara, Christoph Berger, Sung~Min Ha,
  Martin Rozycki, et~al.
\newblock Identifying the best machine learning algorithms for brain tumor
  segmentation, progression assessment, and overall survival prediction in the
  brats challenge.
\newblock {\em arXiv preprint arXiv:1811.02629}, 2018.

\bibitem{baur2021autoencoders}
Christoph Baur, Stefan Denner, Benedikt Wiestler, Nassir Navab, and Shadi
  Albarqouni.
\newblock Autoencoders for unsupervised anomaly segmentation in brain mr
  images: A comparative study.
\newblock {\em Medical Image Analysis}, page 101952, 2021.

\bibitem{baur2018deep}
Christoph Baur, Benedikt Wiestler, Shadi Albarqouni, and Nassir Navab.
\newblock Deep autoencoding models for unsupervised anomaly segmentation in
  brain mr images.
\newblock In {\em International MICCAI Brainlesion Workshop}, pages 161--169.
  Springer, 2018.

\bibitem{berg2019unsupervised}
Amanda Berg, J{\"o}rgen Ahlberg, and Michael Felsberg.
\newblock Unsupervised learning of anomaly detection from contaminated image
  data using simultaneous encoder training.
\newblock {\em arXiv preprint arXiv:1905.11034}, 2019.

\bibitem{bilic2019liver}
Patrick Bilic, Patrick~Ferdinand Christ, Eugene Vorontsov, Grzegorz Chlebus,
  et~al.
\newblock The liver tumor segmentation benchmark (lits).
\newblock {\em arXiv:1901.04056}, 2019.

\bibitem{carass2017longitudinal}
Aaron Carass, Snehashis Roy, Amod Jog, Jennifer~L Cuzzocreo, Elizabeth Magrath,
  Adrian Gherman, Julia Button, James Nguyen, Ferran Prados, Carole~H Sudre,
  et~al.
\newblock Longitudinal multiple sclerosis lesion segmentation: resource and
  challenge.
\newblock {\em NeuroImage}, 148:77--102, 2017.

\bibitem{chen2018unsupervised}
Xiaoran Chen and Ender Konukoglu.
\newblock Unsupervised detection of lesions in brain mri using constrained
  adversarial auto-encoders.
\newblock {\em arXiv preprint arXiv:1806.04972}, 2018.

\bibitem{chlebus2018automatic}
Grzegorz Chlebus, Andrea Schenk, Jan~Hendrik Moltz, Bram van Ginneken,
  Horst~Karl Hahn, and Hans Meine.
\newblock Automatic liver tumor segmentation in ct with fully convolutional
  neural networks and object-based postprocessing.
\newblock {\em Scientific reports}, 8(1):1--7, 2018.

\bibitem{del2016discriminative}
Allison Del~Giorno, J~Andrew Bagnell, and Martial Hebert.
\newblock A discriminative framework for anomaly detection in large videos.
\newblock In {\em European Conference on Computer Vision}, pages 334--349.
  Springer, 2016.

\bibitem{dey2018compnet}
Raunak Dey and Yi~Hong.
\newblock Compnet: Complementary segmentation network for brain mri extraction.
\newblock In {\em International Conference on Medical Image Computing and
  Computer-Assisted Intervention}, pages 628--636. Springer, 2018.

\bibitem{dey2020hybrid}
Raunak Dey and Yi~Hong.
\newblock Hybrid cascaded neural network for liver lesion segmentation.
\newblock In {\em 2020 IEEE 17th International Symposium on Biomedical Imaging
  (ISBI)}, pages 1173--1177. IEEE, 2020.

\bibitem{erfani2016high}
Sarah~M Erfani, Sutharshan Rajasegarar, Shanika Karunasekera, and Christopher
  Leckie.
\newblock High-dimensional and large-scale anomaly detection using a linear
  one-class svm with deep learning.
\newblock {\em Pattern Recognition}, 58:121--134, 2016.

\bibitem{giordana1997estimation}
Nathalie Giordana and Wojciech Pieczynski.
\newblock Estimation of generalized multisensor hidden markov chains and
  unsupervised image segmentation.
\newblock {\em IEEE Transactions on Pattern Analysis and Machine Intelligence},
  19(5):465--475, 1997.

\bibitem{goodfellow2014generative}
Ian Goodfellow, Jean Pouget-Abadie, Mehdi Mirza, Bing Xu, David Warde-Farley,
  Sherjil Ozair, Aaron Courville, and Yoshua Bengio.
\newblock Generative adversarial nets.
\newblock In {\em Advances in neural information processing systems}, pages
  2672--2680, 2014.

\bibitem{kimura2018anomaly}
Masanari Kimura and Takashi Yanagihara.
\newblock Anomaly detection using gans for visual inspection in noisy training
  data.
\newblock In {\em Asian Conference on Computer Vision}, pages 373--385.
  Springer, 2018.

\bibitem{lee2002unsupervised}
Te-Won Lee and Michael~S Lewicki.
\newblock Unsupervised image classification, segmentation, and enhancement
  using ica mixture models.
\newblock {\em IEEE Transactions on Image Processing}, 11(3):270--279, 2002.

\bibitem{menze2014multimodal}
Bjoern~H Menze, Andras Jakab, Stefan Bauer, Jayashree Kalpathy-Cramer, Keyvan
  Farahani, Justin Kirby, Yuliya Burren, Nicole Porz, Johannes Slotboom, Roland
  Wiest, et~al.
\newblock The multimodal brain tumor image segmentation benchmark (brats).
\newblock {\em IEEE transactions on medical imaging}, 34(10):1993--2024, 2014.

\bibitem{o2004combined}
Robert~J O'Callaghan and David~R Bull.
\newblock Combined morphological-spectral unsupervised image segmentation.
\newblock {\em IEEE transactions on image processing}, 14(1):49--62, 2004.

\bibitem{puzicha1999histogram}
Jan Puzicha, Thomas Hofmann, and Joachim~M Buhmann.
\newblock Histogram clustering for unsupervised image segmentation.
\newblock In {\em Proceedings. 1999 IEEE Computer Society Conference on
  Computer Vision and Pattern Recognition (Cat. No PR00149)}, volume~2, pages
  602--608. IEEE, 1999.

\bibitem{ronneberger2015u}
Olaf Ronneberger, Philipp Fischer, and Thomas Brox.
\newblock U-net: Convolutional networks for biomedical image segmentation.
\newblock In {\em International Conference on Medical image computing and
  computer-assisted intervention}, pages 234--241. Springer, 2015.

\bibitem{rubner2000earth}
Yossi Rubner, Carlo Tomasi, and Leonidas~J Guibas.
\newblock The earth mover's distance as a metric for image retrieval.
\newblock {\em International journal of computer vision}, 40(2):99--121, 2000.

\bibitem{schlegl2019f}
Thomas Schlegl, Philipp Seeb{\"o}ck, Sebastian~M Waldstein, Georg Langs, and
  Ursula Schmidt-Erfurth.
\newblock f-anogan: Fast unsupervised anomaly detection with generative
  adversarial networks.
\newblock {\em Medical image analysis}, 54:30--44, 2019.

\bibitem{schlegl2017unsupervised}
Thomas Schlegl, Philipp Seeb{\"o}ck, Sebastian~M Waldstein, Ursula
  Schmidt-Erfurth, and Georg Langs.
\newblock Unsupervised anomaly detection with generative adversarial networks
  to guide marker discovery.
\newblock In {\em International conference on information processing in medical
  imaging}, pages 146--157. Springer, 2017.

\bibitem{seebock2016identifying}
Philipp Seeb{\"o}ck, Sebastian Waldstein, Sophie Klimscha, Bianca~S Gerendas,
  Ren{\'e} Donner, Thomas Schlegl, Ursula Schmidt-Erfurth, and Georg Langs.
\newblock Identifying and categorizing anomalies in retinal imaging data.
\newblock {\em arXiv preprint arXiv:1612.00686}, 2016.

\bibitem{shi2000normalized}
Jianbo Shi and Jitendra Malik.
\newblock Normalized cuts and image segmentation.
\newblock {\em IEEE Transactions on pattern analysis and machine intelligence},
  22(8):888--905, 2000.

\bibitem{zenati2018efficient}
Houssam Zenati, Chuan~Sheng Foo, Bruno Lecouat, Gaurav Manek, and
  Vijay~Ramaseshan Chandrasekhar.
\newblock Efficient gan-based anomaly detection.
\newblock {\em arXiv preprint arXiv:1802.06222}, 2018.

\bibitem{zenati2018adversarially}
Houssam Zenati, Manon Romain, Chuan-Sheng Foo, Bruno Lecouat, and Vijay
  Chandrasekhar.
\newblock Adversarially learned anomaly detection.
\newblock In {\em 2018 IEEE International Conference on Data Mining (ICDM)},
  pages 727--736. IEEE, 2018.

\end{thebibliography}
}
\end{document}